# Designing an FPGA Synthesizable Computer Vision Algorithm to Detect the Greening of Potatoes

Jaspinder Pal Singh
*Assistant Professor Electronics and Communication Engineering*
*Guru Nanak Institute of Engineering and Management*
*Hoshiarpur, Punjab, India.*

*Abstract--* **Potato quality control has improved in the last years thanks to automation techniques like machine vision, mainly making the classification task between different quality degrees faster, safer and less subjective. In our study we are going to design a computer vision algorithm for grading of potatoes according to the greening of the surface colour of potato. The ratio of green pixels to the total number of pixels of the potato surface is found. The higher the ratio the worse is the potato. First the image is converted into serial data and then processing is done in RGB colour space. Green part of the potato is also shown by de-serializing the output. The same algorithm is then synthesized on FPGA and the result shows thousand times speed improvement in case of hardware synthesis.**

## I. INTRODUCTION

Computer vision is a field that includes methods for acquiring, processing, analyzing, and understanding images and, in general, high-dimensional data from the real world in order to produce numerical or symbolic information. Computer vision has also been described as the enterprise of automating and integrating a wide range of processes and representations for vision perception. Computer vision has become an essential technology for quality control in the food industry, which continuously demands new and better applications.

*A. Potato*

The potato (*Solanum tuberosum*) is an herbaceous annual that grows up to 100 cm (40 inches) tall and produces a tuber also called potato so rich in starch that it ranks as the world's fourth most important food crop, after maize, wheat and rice. The major markets for raw and processed potatoes are grocery stores, fast-food chains, restaurants, and potato chip makers. Before the storage of potatoes in cold stores, separating of infected tubers from healthy ones is necessary. Potato grading has been determined as one of the qualitative factors which are considered by sellers and purchasers. Grading is usually done by workers and observing the tuber surface. Potato grading by workers include some disadvantages such as lack of stability, time and cost consuming.

*B. Potato blemishes*

There are a number of diseases affecting potato tubers that generally of little or no health consequence to humans, strongly and negatively influence consumer choice. These include black dot, silver scurf, powdery scab, common scab, and skin spot. The fungal species of Rhizoctonia Solani also causes significant skin blemish as black scurf and elephant hide. Other forms of blemish include physical damage, e.g. growth cracks, mechanical damage and slug damage as well as physiological blemishes, e.g. greening and sprouting. Figure no.1 shows some of the common potato blemishes like scurf's, black dots, greening, scab etc. These all are the surface defects of the potatoes and can be detected with the help of machine vision systems. In typical machine vision systems for quality analysis of food products, there are several major steps: after pre-processing (e.g. to segment the object of interest from the background), image features are extracted that summarise important qualities of the object, then a pattern recognition system is used to categorise the input data.

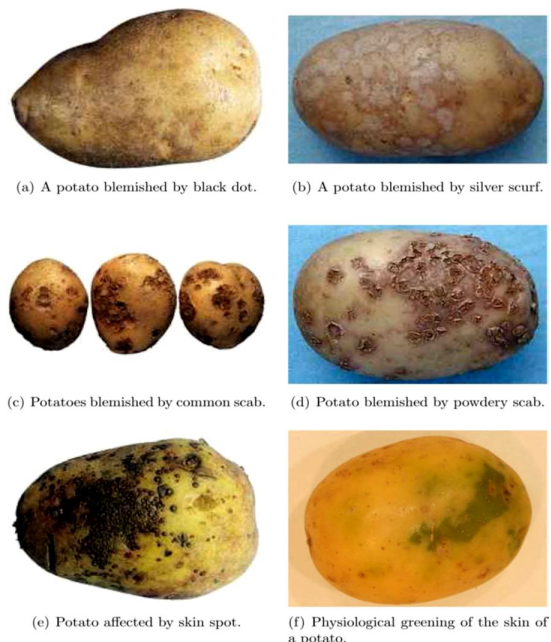

(a) A potato blemished by black dot.  (b) A potato blemished by silver scurf.
(c) Potatoes blemished by common scab.  (d) Potato blemished by powdery scab.
(e) Potato affected by skin spot.  (f) Physiological greening of the skin of a potato.

Fig. 1 Common Potato Blemishes

*C. USDA standards for potato greening*

According to the USDA standards for potato grading [1] a potato is said to be damaged when removal of green part causes a loss of more than 5 percent of the total weight of the potato or when green colour affects more than 25 percent of the surface in the aggregate. Likewise a potato is said to be seriously damaged when removal of the green part causes a loss of more than 10 percent of the weight of the potato or when green colour affects more than 50 percent of the surface in the aggregate. In our work we will consider the factor of green surface aggregate and not the potato weight since we cannot calculate the weight of green part using image processing.

*D. SIMULINK HDL Coder*

HDL Coder generates portable, synthesizable VERILOG and VHDL code from MATLAB functions, SIMULINK models, and State flow charts. The generated HDL code can be used for FPGA programming or ASIC prototyping and design.

HDL Coder provides a workflow advisor that automates the programming of XILINX and Altera FPGAs. You can control HDL architecture and implementation, highlight critical paths, and generate hardware resource utilization estimates. HDL Coder provides traceability between your SIMULINK model and the generated VERILOG and VHDL code, enabling code verification for high-integrity applications adhering to DO-254 and other standards.

*E. Related work*

To increase the performance and speed of image processing algorithms Danny crookes [2] discussed four different hardware applicable approaches. He suggested that among the four approaches, dynamically reprogrammable hardware (FPGAs) offers the potential of sufficiently high performance at an acceptably low cost, for certain applications including image processing. Unay and Gosselin [3] developed methods to distinguish between blemishes in apples and healthy apples with visible stem or calyx. Images were recorded using special filters to restrict the observed light frequencies, and then various features including statistical moments and shape features were used for pattern recognition. Bolle et al. [4] developed the Veggie-Vision system, using HSV-colour and texture histograms to classify different types of fruit and vegetables, with application to a supermarket check-out for automatic produce recognition. Jelinski et al. [5] introduced visual inspection methods for pasteurised cheese. They also used thresholding to detect ingredients such as chives, and developed methods to measure the distribution and quantity of the detected ingredients. Munkevik et al. [6] developed a machine vision system for automatic descriptive sensory evaluation of meals, where a neural network was trained to mimic the opinion of human experts in describing the sensory attributes of a prototypical meal. Jarimopas and Jaisin [6] introduced a system for sorting sweet tamarind, by measuring the size and shape of tamarind pods as well as detecting defects in the form of broken pods. Threshold intensity values were used to distinguish blemishes from non-blemishes.

Recently, lot of research have been carried out in the field of agriculture produce grading by depending on the computer, in order to reduce the processing time and to provide accurate results. Digital image processing, as a computer based technique, has been extremely used by scientists to solve problems in agriculture [7].

In the case of potato product, most of the research works focused on potato inspection without singulation [8][9] and blemish detection like black dot, silver surface, common scab, etc. [10]. In the area of machine vision for potatoes, Tao et al. [11] used Fourier harmonics to describe the shapes of potatoes, forming a metric based on the first 10 Fourier harmonics of the potato's outline to develop a classification method which agreed with human classification 89.2% of the time. Muir et al. [12] used custom lighting equipment to project light at a variety of different wavelengths to demonstrate the different reflective properties of specific blemishes at each wavelength. Tao et al. [13] describes the use of the HSI colour space for identifying greened potatoes as well as yellow and green apples. This was done by use of histograms produced from each of the HSI channels. It was noted that more bins in a histogram resulted in a higher performance. Heinemann et al. [14] graded potatoes by size and shape to meet United States Department of Agriculture (USDA) standards. Size was measured by the longest distance between two points on the boundary, while the shape was determined using Fourier descriptors. The system achieved 97–98% accuracy when classifying stationary potatoes but dropped to between 77% and 88% when tested on moving potatoes. Jing Jin [15] proposed a novel inspection approach to detect external defects of potatoes consisting of AII or FII method, Otsu method, morphological operation, feature extraction, and defect recognition. FII is a simple and easy method which relies on uniform illumination, while AII can deal with both uniform and non-uniform illumination. The combination of Otsu method and morphological operation is capable of segmenting the suspect defects and extracting structure features such as colour and area. Zhou et al. [16] developed a system using green levels to detect green defects (greening and sprouting) in individual potatoes. They also classified potatoes in terms of shape by comparison to an ellipse template and in size and weight by measuring the minor axis and area, respectively. Guannan et al. [17] detected misshapen potatoes by comparing the local rate of change of the radius of a potato. In addition they detected sprouting using a comparison of the green colour channel with the intensity. This value at each pixel was compared to the average value across the potato and if the difference was above a threshold the pixel was determined to be part of a sprout.

The two main issues associated with potato greening are human health and marketability. Human health is a concern because of the independent and parallel development of steroidal glycoalkaloids in green tubers [18]. Glycoalkaloids are a naturally occurring and toxic group of secondary plant compounds found throughout the foliage and tubers of members of the Solanaceae. In small amounts, glycoalkaloids contribute to potato flavour. However, at higher levels, glycoalkaloid consumption can result in symptoms ranging from nausea to coma and even death in extreme cases [19].

Maleeha Kiran et al [20] investigated the performance of an automated surveillance system. The reduction in processing speed can be achieved if a component of the surveillance system is embedded onto a hardware based platform like an FPGA. The results obtained indicated that the processing speed of the component was constantly faster on the FPGA platform as compared to MATLAB or C++ environment.

.Potatoes contains toxic compounds known as *glycoalkaloids*, of which the most prevalent are *solanine* and *chaconine*. This toxin affects the nervous system, causing weakness and confusion. Exposure to light, physical damage, and age increase *glycoalkaloid* content within the tuber. The highest concentrations occur just underneath the skin. Light exposure causes greening from chlorophyll synthesis, thus giving a visual clue as to areas of the tuber that may have become more toxic. Since consumption of green parts of potato is harmful to human, this is very important to develop an inspection system to reject green potatoes during sorting process. Hence, this paper aims to introduce a machine vision algorithm to estimate the potato green surface area.

## II. MATLAB IMPLEMENTATION

### A. Image Acquisition

The processing of the coloured image of potato is done in MATLAB. Images of dimension 640*480 of nine defected green surface potatoes are captured as shown in figure no.2. Each image uses nearly 50-60 KB of memory for storage. The background of the image should be kept white as choosing any other background colour will lead to the inefficiency of our algorithm. The distance between the lens and potato is fixed (nearly 50 cm).

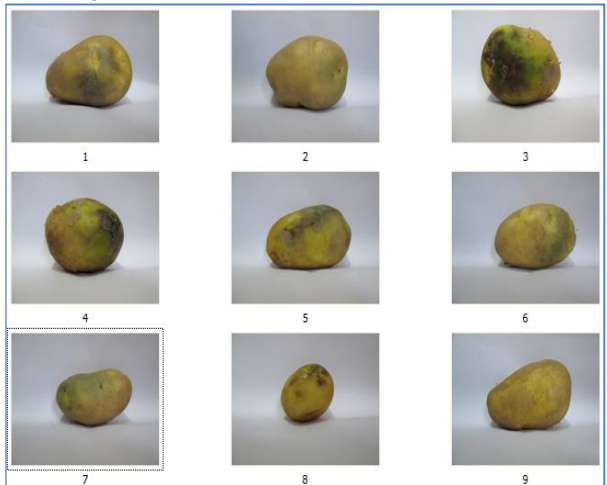

Fig. 2 Sample Space of images of green surface potatoes

Also some artificial lighting was done in order to get the proper colour of the surface of potato in the image. Flash of the camera should be kept off as it may lead to some loss of data in the image. There should be minimum shadow of the potato present in the image as our algorithm may include the shadow part of the potato inside the ROI which will lead to inefficient result. Figure no.3 shows one of the images taken from the sample space on which we will apply our algorithm. Green part of the potato surface is encircled in the image.

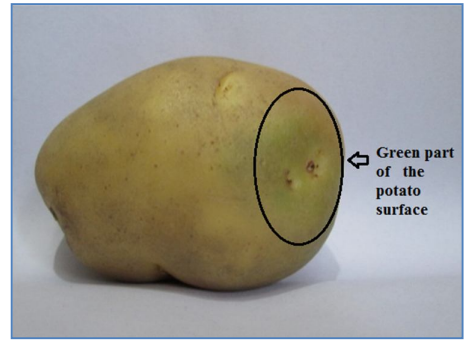

Fig. 3 Image showing green surface on Potato

### B. Feature Extraction

An algorithm is designed in MATLAB for the detection of green colour on potato surface using the R, G and B values of the image. The algorithm is then embedded in the MATLAB SIMULINK model that uses the algorithm for grading of potatoes according to the extent of greening of the surface.

In the SIMULINK model as shown in figure no.4 various pre-defined and user-defined blocks are used. The R, G and B component of the image is first separated and then the data is serialized using various Simulink blocks. Simultaneously the same data is send to another function block which computes the total number of pixels present in the ROI (region of interest) and also displays the ROI.

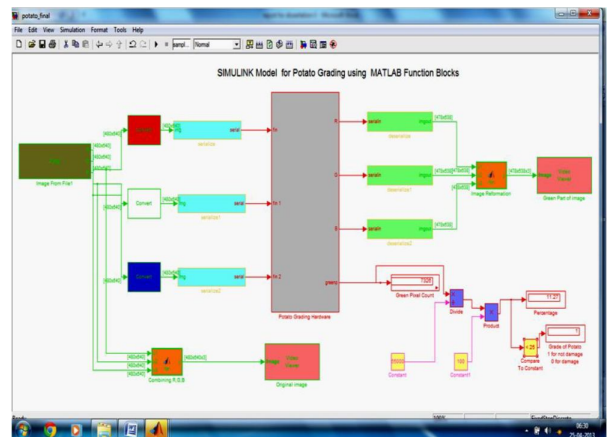

Fig. 4 SIMULINK model

The ROI includes the area of the image where potato is present, hence eliminating the white background. It is very important to first obtain the ROI because ultimately we have to compute the ratio of greener pixels to total number of pixels present in the ROI and to obtain the latter we first have to find the area falling under the ROI. ROI is computed by thresholding the value of B component of the image. The figure no.5 shows the region of interest present in figure no.3. It is important to note here that no shadow of the potato should be present in the image otherwise it will also be included in the ROI and also the background of the image should be white.

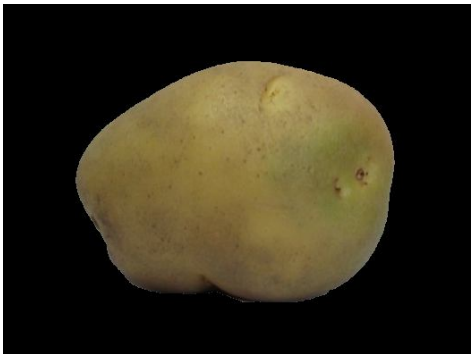

Fig. 5 Region of interest (ROI)

The computation of green pixels is done by thresholding the difference between the R and G values of the image. For greener part of potato surface this difference is less as compared to normal (non-greener surface). After calculating the total number of green pixels in the ROI, the ratio of green pixels to the total number of pixel in the ROI is found and the percentage is calculated as shown in figure no.6. The higher the percentage value, the worse is the potato.

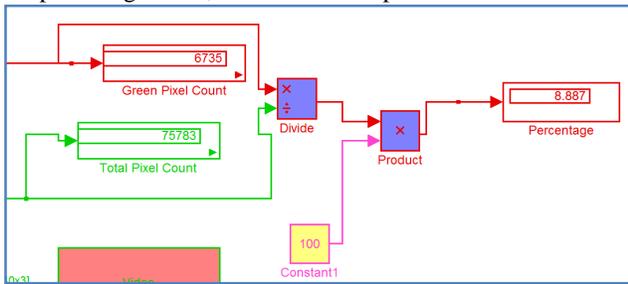

Fig. 6 Percentage Value of Green part

After the execution of the whole process an image is also displayed which shows the green part present on the potato surface as shown in figure no.7

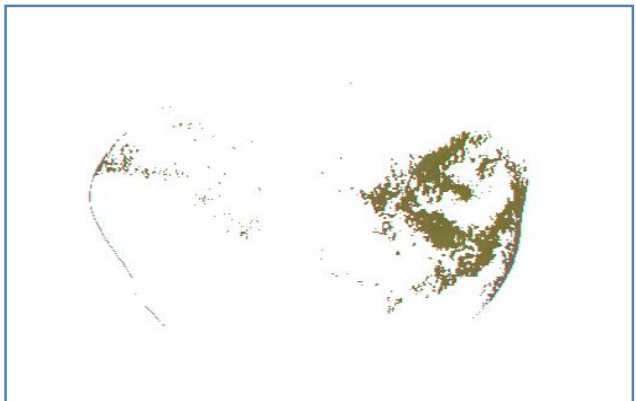

Fig. 7 Green part displayed by software

After the execution of algorithm in MATLAB, a VHDL code is generated using SIMULINK HDL Coder

### C. VHDL Code and Hardware synthesis

HDL Workflow Advisor is available in the tools menu of the SIMULINK. List of processes are needed to be executed sequentially before the Workflow Advisor generates the VHDL code for the Matlab Algorithm. It is very important to note here that it is not possible to convert every Matlab algorithm into VHDL code. The Algorithm should be written in such a way that it can be easily implemented on hardware otherwise the tool will not generate the code. The HDL workflow advisor generates six different VHDL files and all of them are embedded in a single project in XILINX ISE 12.3. The code is synthesized on SPARTAN 3E family, device XC3S250E, package PQ208 and speed -5.

### D. Simulink Profiler Report and Synthesis Report

After the simulation of code in MATLAB, a SIMULINK profiler report is generated which shows the timing constraints. The processing is done pixel by pixel. As the total number of pixels is 640*480=307200, the function is called or processed 307200 times. Similarly when the VHDL code is synthesized on FPGA, timing report is generated by the Xilinx Software. Figure no. 8 shows the screenshot of the timing reports of software simulation and hardware synthesis. The timing details are also written in table I and II on the next page.

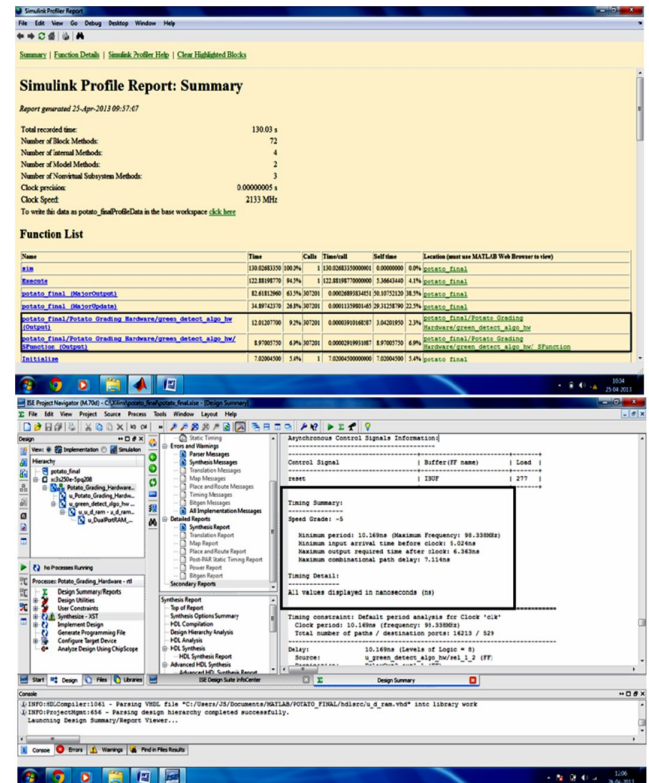

Fig. 8 Timing report of Simulink and Xilinx

### III. CONCLUSION AND FURTHER WORK

An algorithm for the detection of green colour on the surface of potato is designed in MATLAB. As the detection of green colour is done in the RGB colour space the algorithm is less time consuming and efficient. The same algorithm is run on two different platforms, one is the software implementation and other is the hardware synthesis. Timing report is generated by MATLAB as well as XILINX and is shown in table I and table II respectively. It is important to note here that the time delays shown in table II is per one pixel i.e. the time consumed by the FPGA to process one pixel. The value shown in the respective table should be multiplied by 307200 to get the processing time for the whole image. Hence, it is clear by reading data from both the tables that the hardware synthesis is producing the output nearly a thousand times faster as compared to the software implementation. Hence the hardware approach works faster as compared to the software approach.

TABLE I
TIMING REPORT OF MATLAB IMPLEMENTATION

| Function name | Time (sec) | Calls | Time/call (sec) |
|---|---|---|---|
| Potato_final/potato_grading_hardware/ green_detect_algo_hw (output) | 12.0120 7700 | 307201 | 0.00003910 168587 |

TABLE II
TIMING REPORT OF HARDWARE SYNTHESIS

| Sr. no. | Timing Constraint | Delay (in nano seconds) |
|---|---|---|
| 1) | Minimum period | 10.169ns |
| 2) | Minimum input arrival time before clock | 9.024ns |
| 3) | Maximum output required time after clock | 6.363ns |
| 4) | Maximum combinational path delay | 7.114ns |

However there is some inefficiency in the case of selection of ROI because if the shadow of potato is present in the image the algorithm will include it in the ROI and this will mislead the results. So special care should be taken that there should be no shadow of the potato present in the image.

In my work the grading of the potatoes is not done because the image of the potato surface is taken from one angle only. It does not cover the whole surface of potato. So, if we want to grade the potatoes according to the USDA standards as discussed earlier, we should analyze the whole surface of the potato. So in further work we can capture the images from different directions so that we can analyze the whole surface of potato and be able to grade the potatoes according to the USDA standard. We can also detect the other potato blemishes like black dots, silver scurf, scab, powdery scab, skin spots etc using image processing and implement it on hardware using VHDL.

ACKNOWLEDGMENT